\documentclass[a4paper]{article}
\pdfoutput=1
\usepackage{INTERSPEECH2022}
\usepackage{latexsym}
\usepackage{amsmath}
\usepackage{url}
\usepackage{amssymb}
\usepackage{amsfonts}
\usepackage{graphicx}
\usepackage{tabularx}
\usepackage{multirow}
\usepackage{arydshln}
\usepackage{mathtools,nccmath}
\usepackage[T5]{fontenc}
\usepackage{enumitem}
\usepackage{todonotes}
\usepackage[hidelinks]{hyperref}

\title{A High-Quality and Large-Scale Dataset \\ for English-Vietnamese Speech Translation}

\name{Linh The Nguyen$^\ast$, Nguyen Luong Tran$^\ast$, Long Doan\sthanks{\ \ The first three authors contributed equally to this work.} , Manh Luong, Dat Quoc Nguyen}
\address{VinAI Research, Hanoi, Vietnam}
\email{\{v.linhnt140, v.nguyentl12, v.longdct, v.manhlt3, v.datnq9\}@vinai.io}

\begin{document}
\maketitle

\begin{abstract}
In this paper, we introduce a high-quality and large-scale benchmark dataset for English-Vietnamese speech translation  with 508 audio hours, consisting of 331K triplets of (sentence-lengthed audio, English source transcript sentence, Vietnamese target subtitle sentence). We also conduct empirical experiments using strong baselines and find that the traditional ``Cascaded'' approach still outperforms the modern ``End-to-End'' approach. To the best of our knowledge, this is the first large-scale English-Vietnamese speech translation study.  We hope both our publicly available dataset and study can serve as a starting point for future research and applications on English-Vietnamese speech translation. 
\end{abstract}

\medskip


\section{Introduction}
\label{sec:intro}


Speech translation---the task of translating speech in one language typically to text in another---has attracted interest for many years \cite{sperber-paulik-2020-speech,iwslt2020,iwslt2021}. However, the development of speech translation systems has been largely limited to the high-resource language pairs because most public available datasets for speech translation are exclusively for the high-resource languages \cite{bilingual-jp-en, fisher-spanish, iwslt2018, libri-french,bstc, mustc,covost1,europarl,tedx}. For example, bilingual speech translation datasets \cite{bilingual-jp-en, fisher-spanish, iwslt2018, libri-french,bstc} are all large-scale ones with about 200+ audio hours, which offer translations either from English to a target language (e.g. French, German and Japanese) or from a source language (e.g. Chinese, Japanese and Spanish) to English. The one-to-many dataset MuST-C~\cite{mustc} is created to facilitate the training of speech translation systems from the source language English into 14 target languages. On the contrary, the many-to-one dataset CoVoST~\cite{covost1} provides speech translations from 11 languages into English.  EuroParl-ST~\cite{europarl} is a many-to-many speech translation dataset that provides translations between 6 European languages, with a total of 30 translation directions, using the debates held in the European Parliament from 2008 to 2012. The Multilingual TEDx
corpus ~\cite{tedx} is created from TEDx talks in 8 source languages with their manual transcriptions and translations in 6 target languages. 

From a societal, linguistic, machine learning, cultural and cognitive perspective \cite{donlpotherlanguages}, it is also worth investigating the speech translation task for low-resource languages, e.g. Vietnamese with about 100M speakers \cite{Ethnologue}. Vietnam is now an attractive destination for trade, investment and tourism \cite{vietnamworkingpaper}. Thus the demand for high-quality speech translation from the global language English to Vietnamese has rapidly increased. To our best knowledge, there is no existing research work that focuses solely on speech translation to Vietnamese. The only available resource for speech translation to Vietnamese is the 441-hour English-Vietnamese speech translation data from  MuST-C \cite{mustc}. However, MuST-C is a TED-talk-based multilingual dataset, thus its authors could not pay attention to any specific language pairs. Thus it might contain  timestamp alignment errors between English audio-transcript pairs and misalignment between English-Vietnamese source-target sentence pairs. 

We use a total of 50 working hours to manually inspect both the validation set (including 1350 triplets with 2.5 audio hours in total) and the test set (including 2361 triplets with 4.1 audio hours) from the MuST-C English-Vietnamese  data. Here, each triplet is checked by two out of our first three authors independently. After cross-checking, we find that: \textbf{5.63\%} of the validation set and \textbf{4.10\%} of the test set have an incorrect audio start or end timestamp of an English source sentence; \textbf{16.15\%} of the validation set and \textbf{9.36\%} of the test set have misaligned English-Vietnamese sentence pairs  (i.e. completely different sentence meaning or partly preserving the meaning). Note that performing a similar manual inspection on the MuST-C English-Vietnamese  training set would take $(441-2.5-4.1)/(2.5+4.1)\times50 = 3290$ working hours, beyond our current human resource. As training/validation/test split is random, the substantial rates of incorrect timestamps and misalignment on the validation and test sets imply that the MuST-C English-Vietnamese  training set might also not reach a high-quality level. 

Our work is to help tackle that issue of low-quality data for English-Vietnamese speech translation. Our contributions are summarized as follows:

\begin{itemize}
    \item As the {first contribution}, we present a high-quality and large-scale English-Vietnamese speech translation dataset containing 508 audio hours.  We hope that our dataset construction process can be further adapted to create more speech translation data for other low-resource languages.
    
    \item As the {second contribution}, on our dataset, we empirically investigate strong baselines to  compare traditional ``Cascaded'' and modern ``End-to-End'' approaches. We find that the  ``Cascaded'' approach still does better than the ``End-to-End'' approach. To the best of our knowledge, this is the first large-scale empirical study for English-Vietnamese speech translation. 
    
    \item As our {final contribution},  we publicly release our dataset for non-commercial use under the CC BY-NC-ND 4.0 license at: \url{https://github.com/VinAIResearch/PhoST}.  We hope both our dataset and empirical study can serve as a starting point for future English-Vietnamese speech translation research and applications. 
\end{itemize}

\section{Our dataset}
\label{sec:phost_dataset}

Our dataset construction process consists of five phases, as detailed in the following subsections.

\subsection{Collecting audio files and transcripts} 
\label{ssec:collect}

We employ the TED2020-v1 corpus  \cite{dataset_ted} of 3125 parallel  English-Vietnamese transcript-subtitle pairs of TED talks. Here, the English transcript document (i.e. source document) was human-translated to the Vietnamese subtitle document (i.e. target document). 
For each talk, we download the corresponding video with the highest quality version and extract the audio from the downloaded video. Here, we extract the audio from each downloaded video by using the FFMPEG library.\footnote{\url{http://ffmpeg.org/}} All extracted audios are formatted as 16kHz WAV files. There are also 5 talks without publicly available videos. We thus obtain 3125 - 5 = 3120 TED-talk-based triplets of (audio file, English transcript document, Vietnamese subtitle document).

\subsection{Pre-processing and sentence segmentation} \label{ssec:preprocess}

We manually check the obtained triplets and find that there are 23 non-English audio files of TED talks, i.e. 23 English source documents are subtitles of these TED talks, not transcripts. There are also 10 audio files using almost all of their time to display songs. We thus remove those 23 + 10 = 33 triplets, resulting in  3120 - 33 = 3087 remaining triplets. To extract sentences for parallel audio-sentence alignments in next phases, we perform sentence segmentation by using Stanford CoreNLP \cite{corenlp} for the English documents and VnCoreNLP \cite{NguyenNVDJ2018,vncorenlp} for the Vietnamese documents.

Note that TED-talk transcripts and subtitles contain non-speech artifacts of audience-related information marked with parentheses (e.g. ``(applause)'', ``(laugh)'' and the like) and speaker identity. We thus remove all of the audience-related information. For removing the speaker identity, we use an effective heuristic of removing words/phrases that start at the beginning of a sentence, consist of at most 10 characters or 3 words, and are followed by the punctuation mark colon (here, the colon is also removed together with the speaker identity).

\subsection{Extracting the audio start and end timestamps for each English sentence} \label{ssec:au-en}

From a pair of TED-talk audio and its English transcript document, to extract the start and the end timestamps from the audio for each sentence in the document, following the MuST-C process \cite{mustc}, we employ the Gentle forced aligner \cite{gentle} that is based on the Kaldi ASR toolkit \cite{kaldi}. 
The Gentle forced aligner takes the audio and transcript document pair as input and outputs a timestamp for each word token in the document. We use the timestamp of the first word in each sentence as the start timestamp of the sentence. We determine the end timestamp of the sentence as $\min(x + 0.5 \text{\ second}, y - 0.01 \text{\ second})$ in which $x$ is the timestamp of the last word in the sentence while $y$ is the timestamp of the first word in the next sentence. 

Note that there are 10K English sentences where the Gentle forced aligner cannot detect a timestamp of the first or the last word in a sentence due to the corresponding background noises (in most cases), multiple-speaker overlaps or non-English sounds in the audio file. We thus manually correct the start and the end timestamps of these sentences.  Here, we develop a small tool based on the PyDub library, which is to play a TED talk audio and to display the audio timestamps, and also to play an audio span given the audio and a start and end timestamp pair.\footnote{\url{https://github.com/jiaaro/pydub}} Given a sentence that we have to determine the missing timestamps of its first and/or last words, we first locate an audio span in the corresponding TED talk audio, which covers the sentence's part with timestamp-detected words. Through listening to the audio (via using the tool) and looking at the transcript sentence, we then expand the audio span to exactly match the whole given sentence, thus extracting the timestamps of its first and/or last words. 

We divide those 10K sentences into three equal parts. Each of the first three authors manually corrects one of these parts. After that, the fourth author verifies each sentence and makes further revisions if needed. This manual correction process takes a total of 210 working hours from four authors.

\subsection{Aligning parallel English-Vietnamese sentence pairs} \label{ssec:en-vi}

To align parallel sentences within a parallel English-Vietnamese document pair,  following \cite{PhoMT}, we first use Google Translate to translate English source sentences into Vietnamese.\footnote{We use the English-to-Vietnamese translation direction because Google-translating from English to Vietnamese produces better translation than from Vietnamese to English. This is confirmed via BLEU scores in the first two rows in Tables 3 and 4 from \cite{kudo-2018-subword} or human-evaluation results for Google Translate in Table 2 from \cite{PhoMT}.} Then,  to produce parallel English-Vietnamese sentence pairs,  we use three alignment toolkits of  Hunalign \cite{hunalign}, Gargantua \cite{gargantua} and Bleualign \cite{bleualign} to perform an intermediate alignment between the Vietnamese Google-translated versions of the English source sentences and the Vietnamese target sentences. Note that Bleualign only performs alignment between the target sentences and the variants translated into the target language of the source sentences. And in a preliminary experiment, we find that for both Hunalign and Gargantua, performing sentence alignment between the Google-translated variants of the English source sentences and the Vietnamese target sentences produces better quality than performing direct alignment between the English sentences and the Vietnamese sentences. Only sentence pairs, aligned by at least two out of three toolkits, are  selected to be included in the output of this sentence alignment process (here, 99.2\% of both the English and the Vietnamese sentences are included in the output).


\subsection{Post-processing}\label{ssec:post}

From a dataset of 3087 triplets of (TED-talk audio, English transcript document, Vietnamese subtitle document) as shown in Section \ref{ssec:preprocess}, we merge the outputs of the previous two phases to produce 331284 triplets of (sentence-lengthed audio, English source sentence of transcript, Vietnamese target sentence of subtitle). We normalize punctuations in sentence pairs. We split our dataset into training/validation/test sets with a 98.8/0.6/0.6 ratio on the TED-talk level, resulting in 327370 training, 1936 validation and 1978 test triplets at the sentence level.

We manually inspect each triplet in the validation and test sets to qualify our dataset, for checking whether there is: (i) a timestamp misalignment between English audio-transcript pairs,\footnote{For checking the timestamp misalignment, we reuse our PyDub-based tool developed for correcting the timestamps of the first and/or last words of sentences in the third phase (Section \ref{ssec:au-en}).} (ii) a misalignment between English-Vietnamese source-target sentence pairs, and (iii) a low-quality translation from the English source sentence to the Vietnamese target sentence. Each triplet is inspected by two out of the first three authors independently: one inspector checks about $(1936 + 1978) \times 2 / 3 \approx 2609$ triplets in 18 working hours on average.

After cross-checking, we do not find on both validation and test sets any error w.r.t. the start and the end timestamps of each English source sentence. We also find that 3 triplets (\textbf{0.15\%}) on the validation set and 2 triplets (\textbf{0.10\%}) on the test set contain misaligned source-target sentence pairs. In addition, all the remaining source-target sentence pairs on both validation and test sets are at a high-quality translation level. Without any start or end timestamp error and with very small proportions of sentence pair misalignment on the validation and test sets,  we believe that our training set reaches a high-quality standard. Therefore, our dataset obtains a substantially better quality than the TED-talk based MuST-C English-Vietnamese speech translation dataset discussed in Section \ref{sec:intro}.

We remove those 3 + 2 = 5 triplets, resulting in 1933 validation and 1976 test triplets for final use. Table \ref{tab:phost_stats} presents the statistics of our dataset.

\begin{table}[!t]
    \centering
    \caption{Our dataset statistics. ``\#triplets'', ``\#hours'', ``\#en/s'' and ``\#vi/s'' denote the number of triplets, the number of audio hours, the average number of word tokens per English sentence and the average number of syllable tokens per Vietnamese sentence, respectively.}
    \resizebox{7.5cm}{!}{
    \def\arraystretch{1.1}
    \begin{tabular}{l|l|l|l|l}
        \hline
        \textbf{Split} & \textbf{\#triplets} & \textbf{\#hours} & \textbf{\#en/s} & \textbf{\#vi/s} \\
        \hline
        \textbf{Training} & 327370 & 501.59 & 16.55 & 20.94 \\
        \textbf{Validation} & 1933 & 3.13 & 17.24 & 22.22 \\
        \textbf{Test} & 1976 & 3.77 & 19.23 & 25.65 \\
        \hline
    \end{tabular}
    }
    \label{tab:phost_stats}
\end{table}


\section{Speech translation approaches}

On our dataset, we compare two speech translation approaches: \textit{Cascaded} vs. \textit{End-to-End}. 

\subsection{Cascaded} 

The ``Cascaded'' approach combines two main components of English automatic speech recognition (ASR) and English-to-Vietnamese machine translation (MT). For ASR, we train the Fairseq's S2T Transformer model  \cite{fairseq_s2t, s2t_transformer} on our English audio-transcript training data.\footnote{In preliminary experiments, we find that training S2T Transformer produces a slightly better ASR result (i.e. lower word error rate) than both  training Conformer \cite{conformer2,GulatiQCPZYHWZW20} and fine-tuning Wav2Vec2 \cite{wav2vec2}.}  
For MT, we fine-tune the pre-trained sequence-to-sequence model mBART  \cite{mbart} that obtains the state-of-the-art performance for English-Vietnamese machine translation \cite{PhoMT,vinaitranslate}. 

Note that we also perform a data augmentation to extend our MT training data. In particular, we employ the NeMo toolkit \cite{nemo} to perform a normalization process of 2 steps, including: (i) the first step of inverse text normalization is to convert our trained S2T Transformer's automatic ASR output into its written form (e.g. ``in nineteen seventy'' is converted into ``in 1975''), and (ii) the second step is to recover capitalization and punctuation marks (here, the output of the first step is the input for the second step). Given the input of 327370 sentence-lengthed training audios to the trained S2T Transformer, each of steps (i) and (ii) then produces  327370 additional English sentences, thus resulting in an extended training set of 327370 * 3 = 982110 parallel English-Vietnamese sentence pairs for fine-tuning mBART. We also fine-tune mBART on a combination of the 3M-sentence-pair dataset PhoMT \cite{PhoMT} and our extended set (here, we remove PhoMT's pairs that appear in our validation and test sets).

\subsection{End-to-End} 

For an ``End-to-End'' approach that directly translates English speech into Vietnamese text, we study two baselines, including the S2T Transformer model and the UPC's speech translation system Adaptor \cite{gallego2021iwslt} that is the only top performance system at IWSLT 2021 \cite{iwslt2021} with publicly available implementation at the time of our empirical investigation.  
The UPC's Adaptor system employs a Length Adaptor module \cite{lna_finetuning} to adjust representations from the pre-trained Wav2Vec2-based speech encoder \cite{wav2vec2} to the mBART-based language-specific decoder \cite{mbart}. We experiment with the best setting of ``Layer-Norm Attention -- Encoder-Decoder with 2-step Adaptor'' of the UPC's Adaptor.\footnote{\url{https://github.com/mt-upc/iwslt-2021}}

\subsection{Implementation details} \label{ssec:implementation}

In the ``Cascaded'' approach, for validation or test, we also apply the two-step normalization process on the output of the ASR component before feeding into the MT component. We initialize the S2T Transformer model used for ``End-to-End'' speech translation by the S2T Transformer trained for the ASR component in the ``Cascaded'' approach. 


Except for the UPC's Adaptor system that requires the input of raw  WAV audio file, we augment the audio data with SpecAugment \cite{specaugment} and extract the log Mel-filterbank with 80 dimensions as input features. We employ SentencePiece \cite{sentencepiece} to learn a vocabulary of 5K subword types for English and a vocabulary of 8K types for Vietnamese, which are used in the S2T Transformer models. We employ the S2T Transformer  and mBART  implementations from the \texttt{fairseq} library \cite{fairseq}. Here, we train the S2T Transformer models with a configuration of \texttt{s2t-transformer-s}.


We optimize all systems on 4 V100 GPUs (32GB each), using Adam \cite{KingmaB14} with different initial learning rates. In particular, we perform a grid search to select the initial Adam learning rate. The optimal initial learning rates for the ``Cascaded'' ASR component, the ``Cascaded'' MT component, the ``End-to-End'' S2T Transformer and the UPC's Adaptor system are 2e-3, 5e-5, 2e-3 and 4e-4 respectively.

For evaluation, we use beam search with beam size of 5 as decoding algorithm.
Following the IWSLT 2021 evaluation campaign \cite{iwslt2021}, we use the BLEU score as our main evaluation metric in our speech translation experiments. We use SacreBLEU \cite{sacrebleu_metrics} to report the detokenized and case-sensitive BLEU score. 
In our ASR experiments, we use \texttt{fairseq} to calculate WER with lowercased, punctuation removal and using \texttt{13a} tokenizer. 
For both ``Cascaded'' and ``End-to-End'', we choose the model checkpoint that produces the highest BLEU score on the validation set to apply to the test set.

\begin{table}[!t]
\centering
\caption{BLEU scores of each approach on our test set. ``Casc.'' and ``E2E'' denote the ``Cascaded'' and ``End-to-End'' approaches, respectively. (I): mBART fine-tuned on our extended training set. (II): mBART fine-tuned on a combination of the 3M-pair  dataset PhoMT and our extended training set.  \textit{In the ``Cascaded'' approach, the word error rate (WER) computed for the ASR component is 7.06, while BLEU scores of (I) and (II) computed for the text-to-text MT component with ``gold'' English source transcript sentences are 36.48 and 37.41, respectively}. Each score difference between two out of four approaches is statistically significant with p-value $<$ 0.01 based on bootstrap resampling (except the difference between 33.65 and 33.30 is with p-value $<$ 0.05).}
\resizebox{7.5cm}{!}{
\def\arraystretch{1.1}
\begin{tabular}{l|l|l}
\hline
\multicolumn{2}{l|}{\bf Model} & \textbf{BLEU$\uparrow$} \\
\hline
\multirow{2}{*}{\rotatebox[origin=c]{90}{Casc.}} & 
(I) mBART w/ our extended dataset & 33.65 \\
\cline{2-3} & (II) mBART w/ PhoMT combination & \textbf{34.31} \\
\hline
\multirow{2}{*}{\rotatebox[origin=c]{90}{E2E}} & S2T Transformer & 29.98 \\
\cline{2-3} & UPC's Adaptor & \textbf{33.30} \\
\hline

\end{tabular}
}
\label{tab:results}
\end{table}

%

\section{Experimental results}
\label{sec:experiment}

\subsection{Main Results}

Table \ref{tab:results} presents final BLEU score results obtained by each speech translation approach on our test set. For ``Cascaded'' results, with the same automatic ASR output as the model input, it is not surprising that fine-tuning mBART on a combination of PhoMT and our extended training set helps produce a better BLEU score than fine-tuning mBART on only our extended training set (34.31 vs. 33.65), thus showing the effectiveness of a larger training size.  For ``End-to-End'' results,  it is also not surprising that with the ability to leverage strong pre-trained models Wav2Vec2 for ASR and mBART for machine translation, the UPC's Adaptor system obtains a 3+ absolute higher BLEU score than the S2T Transformer model trained solely without pre-training (33.30 vs. 29.98). In a comparison between the ``Cascaded'' and ``End-to-End'' results, we find that the ``Cascaded'' approach does better than the ``End-to-End'' approach. 


Note that our ``Cascaded'' results in Table \ref{tab:results} are reported w.r.t. data augmentation to extend our original training set via the 2-step normalization process. We conduct an ablation study on the validation set to investigate the usefulness of this data augmentation strategy. We find that not using data augmentation leads to a decrease in the BLEU performance score of the ``Cascaded'' approach.  In particular, on the validation set, the ``Cascaded'' approach with mBART fine-tuned on our extended training set (i.e. with data augmentation) obtains a BLEU score of 30.88, while the one with mBART fine-tuned on our original training set (i.e. without data augmentation) obtains a lower BLEU score at 30.18 (i.e. 30.88 $\longrightarrow$ 30.18).

%

\subsection{Comparison with the MuST-C English-Vietnamese}

Recall that sections \ref{sec:intro} and \ref{ssec:post} detail the rates of incorrect timestamps and misalignment on the validation and test sets in the MuST-C English-Vietnamese data and our dataset, respectively; showing that our dataset obtains a substantially better quality than the MuST-C English-Vietnamese data. 

To further investigate the impact of our curation effort, we also conduct experiments comparing model performances on  the  MuST-C English-Vietnamese (En-Vi) training set and our training set. For a fair comparison, as not all of our data are overlapping with MuST-C, we filter from the  MuST-C En-Vi training set all TED talks that appear in our validation and test sets. We then sample a subset of TED talks from our training set so that our sampled subset has the same total number of speech hours as the filtered MuST-C En-Vi training set. We first train two S2T Transformer models for ASR and then two end-to-end Adaptor models for speech translation, using our sampled subset and the filtered MuST-C En-Vi training set. We tune hyper-parameters using our validation set in the same manner as presented in Section \ref{ssec:implementation}. 
We evaluate these models using our test set.  Experimental results obtained for these experiments are presented in Table \ref{tab:curation_experiment}. 

On the ASR task, the S2T Transformer trained using our sampled subset obtains a lower WER than the S2T Transformer trained using the MuST-C data (7.44 vs. 9.09), showing our better audio-transcript alignment. Similarly, on the speech translation task, the end-to-end Adaptor model trained using our sampled subset  does better than the one trained using the MuST-C data (32.37 vs. 31.66), demonstrating the effectiveness of our curation effort w.r.t. the end-to-end setting. Note that when trained using our full training set, as shown in Table \ref{tab:results}, S2T Transformer obtains a lower WER at 7.06 for ASR, while Adaptor obtains a higher BLEU score at 33.30 for speech translation, thus reconfirming the positive impact of a larger training size.

\begin{table}[]
    \centering
    \caption{ASR WER scores and end-to-end speech translation BLEU scores on our test set. Score differences are statistically significant with p-value $<$ 0.01. }
    \label{tab:curation_experiment}
    \resizebox{7.5cm}{!}{
    \def\arraystretch{1.1}
    \begin{tabular}{l|c|c}
        \textbf{Training data} & \textbf{WER$\downarrow$} & \textbf{BLEU$\uparrow$} \\
        \hline
        MuST-C En-Vi training set & 9.09 & 31.66 \\
        Our sampled training subset & 7.44 & 32.37 \\
        \hline
    \end{tabular}
    }
\end{table}

We provide a qualitative example to demonstrate the qualitative differences between the end-to-end Adaptor models trained on MuST-C and on the subset of our dataset. With input audio of the English sentence ``But on a long wavelength sea, you'd be rolling along, relaxed, low energy.'', we have: 

\begin{itemize}
    \item The end-to-end Adaptor model trained on MuST-C provides an output (1) of ``Nhưng trên sóng biển dài, bạn sẽ lăn dọc theo, thư giãn, năng lượng thấp.'' (here, using Google Translate to translate output (1) to English obtains: ``But on the long ocean waves, you'll roll along, relaxed, low energy.'').
    \item The end-to-end Adaptor model trained on the subset of our dataset provides an output (2) of ``Nhưng trên một vùng biển có bước sóng dài, bạn sẽ lăn dọc, thư giãn, ít tốn năng lượng hơn.'' (here, using Google Translate to translate output (2) to English obtains: ``But on a sea with long wavelengths, you will roll along, relax, with less energy.'').
\end{itemize}

\noindent Output (2) is more natural, smooth and better preserving the source sentence's meaning than output (1), thus also implying a higher quality accounted for  our dataset.

\section{Conclusion}

In this paper, we have presented a high-quality and large-scale dataset with 508 audio hours for English-Vietnamese speech translation. On our dataset, we empirically conduct experiments using strong baselines to compare the ``Cascaded'' and ``End-to-End'' approaches. Experimental results show that the ``Cascaded'' approach does better than the ``End-to-End'' approach. We hope that the public release of our  dataset  can serve as the starting point for further English-Vietnamese speech translation  research and applications.

\bibliography{custom}
\bibliographystyle{IEEEtran}

\end{document}